\documentclass[aps,preprint,superscriptaddress, amsmath,amssymb,floatfix,longbibliography]{revtex4-1} %two columns
\usepackage{bm, comment, graphicx, here, amsmath, color, mathrsfs}
\usepackage[normalem]{ulem}
\usepackage[colorlinks=True, allcolors=blue]{hyperref}
\usepackage{balance, setspace, float, titlesec, array}
\usepackage{ulem}
\titlespacing*{\section}{0pt}{1ex}{0ex}
\titlespacing*{\subsection}{0pt}{1ex}{0ex}

\usepackage{xcolor}

\newcommand{\blue}[1]{\textcolor{blue}{#1}}

\usepackage{soul}
\soulregister\cite7
\soulregister\ref7
\soulregister\onlinecite7

\begin{document}
\title{Graph Learning Metallic Glass Discovery from Wikipedia}

\author{K.-C. Ouyang$^\ddagger$}
\affiliation{Songshan Lake Materials Laboratory, Dongguan, China}
\affiliation{Department of Physics, University of Science and Technology of China, Hefei, China}

\author{S.-Y. Zhang$^\ddagger$}
\affiliation{Songshan Lake Materials Laboratory, Dongguan, China}

\author{S.-L. Liu$^\ddagger$}
\affiliation{Songshan Lake Materials Laboratory, Dongguan, China}

\author{J. Tian}
\affiliation{Lawrence Berkeley National Laboratory, Berkeley CA 94720, USA}

\author{Y.-H. Li}
\affiliation{Songshan Lake Materials Laboratory, Dongguan, China}

\author{H. Tong}
\email{huatong@ustc.edu.cn}
\affiliation{Department of Physics, University of Science and Technology of China, Hefei, China}

\author{H.-Y. Bai}
\affiliation{Songshan Lake Materials Laboratory, Dongguan, China}
\affiliation{Institute of Physics, Chinese Academy of Sciences, Beijing, China}

\author{W.-H. Wang}
\affiliation{Songshan Lake Materials Laboratory, Dongguan, China}
\affiliation{Institute of Physics, Chinese Academy of Sciences, Beijing, China}

\author{Y.-C. Hu}
\email{yuanchao.hu@sslab.org.cn \\
$^\ddagger$These authors contributed equally: O.-K., S.-Z. and S.-L.}
\affiliation{Songshan Lake Materials Laboratory, Dongguan, China}

\date{May 15, 2025}

\begin{abstract}
{\bf Synthesizing new materials efficiently is highly demanded in various research fields.
	However, this process is usually slow and expensive, especially for metallic glasses,
	whose formation strongly depends on the optimal combinations of multiple elements
	to resist crystallization. This constraint renders only several thousands of candidates
	explored in the vast material space since 1960. Recently, data-driven approaches
	armed by advanced machine learning techniques provided alternative routes for intelligent materials design. Due to data scarcity and immature material encoding, the
	conventional tabular data is usually mined by statistical learning algorithms, giving
	limited model predictability and generalizability. Here, we propose sophisticated data
	learning from material network representations. The node elements are encoded from
	the Wikipedia by a language model. Graph neural networks with versatile architectures are designed to serve as recommendation systems to explore hidden relationships
	among materials. By employing Wikipedia embeddings from different languages, we
	assess the capability of natural languages in materials design. Our study proposes
	a new paradigm to harvesting new amorphous materials and beyond with artificial
	intelligence.}
\end{abstract}

\maketitle

%--------------------------------------main text----------------------------------

\section*{Introduction}

The past decades have witnessed the fast growth in the number of available metallic materials with a disordered structure, namely metallic glasses (MGs) or amorphous alloys~\cite{inoue_recent_2011,wang_roadmap_2025}. Unlike conventional crystalline alloys, the amorphous nature of MGs endows them with exceptional mechanical, physical and chemical properties. They are not only important prototypical materials to solve critical scientific problems, but also show great potentials in various applications. For instance, they may enable transformative applications ranging from biomedical implants leveraging their biocompatibility to aerospace components utilizing their high strength-to-weight ratios~\cite{Eliaz_biomaterials_2019, Demetriou_damage_2011}. 

One of the most important unresolved problems associated with MGs is their limited glass-forming ability (GFA)~\cite{inoue_stabilization_2000}. 
GFA is usually measured by the critical cooling rate, which is the minimal cooling rate required by a material to bypass crystallization.
Dominated by undirectional metallic bonds, metallic alloys are prone to crystallization with a remarkably faster rate than molecular systems. In experiments, there is always an upper limit of the accessible sample size determined by the critical cooling rate for each composition. So far, the record-keeper is still PdNiCuP discovered in 1997, with a critical diameter of $\sim$ 72 mm~\cite{inoue_preparation_1997}. This severely hinders their wide applications.
How to break the ceiling of this amorphous sample size is urgent to tackle. The critical route is to discover a composition with sluggish crystallization kinetics and enhanced supercooled liquid stability.
There are great endeavors in the past to propose empirical rules to guide MG design. So far, there are nearly 30 physical quantities that have exhibited a certain correlation with GFA for experimentally developed MGs~\cite{ward_machine_2018,liu2020machine}. Unfortunately, they are not agnostic but phenomenological. The common prerequisite is to fabricate the MG. Therefore, the predictive capability is absent.

Trial-and-error experimentation is the main strategy to explore new MGs, with Turnbull's deep eutectic rule~\cite{turnbull_under_1969}, Inoue's three conditions\cite{inoue_stabilization_2000}, and some other experiences as the rule of thumb~\cite{johnson_bulk_1999}. This is rather costly and inefficient considering the combinatorial complexity of elemental interactions and the unknown components. This challenge has motivated the proposal of high-throughput sputtering facility design to enable parallel synthesis of a library of an alloy system with composition gradients in space~\cite{ding_combinatorial_2014, li_high_2019}. This technology remarkably boosts production efficiency but suffers from the high effective cooling rate and trial element combinations. It remains resource-intensive without systematic theoretical guidance. 

Intelligent material design is the new request.
In recent years, data-driven approaches with machine learning algorithms have gained prominence in materials design, including MGs~\cite{sun2017machine,ren_accelerated_2018,ward_machine_2018,wen2019machine,liu2020machine,lu2020interpretable,xiong2020machine,peng2021determination,batra_emerging_2021,afflerbach_machine_2022,li_data_2022,forrest2023evolutionary,merchant_scaling_2023,liu2023machine,liu2024effective,hu_data_2023,xie_catalogue_2025}. By leveraging accumulated experimental and computational datasets, statistical learning methods greatly forwarded the frontlines of MG design. Unprecedented physical insights were unveiled especially from the high-dimensional latent space.
However, the generalization capability of these models is still rather limited, rendering constrained prediction power in new MG design.
There are several inherent reasons. 
Firstly,  the available dataset is rather small and imbalanced.
Secondly, tabular data representation is the common strategy to call for the statistical learning models. The supervised learning frameworks predominantly optimize for known compositional patterns, lacking capacity to infer latent chemical interaction principles governing amorphous phase stability.
Thirdly, the materials are generally encoded by the physical properties of the involved elements and their composition regulations, and some alloy properties as well. The descriptor selection introduces domain knowledge bias, potentially overlooking non-canonical glass-forming mechanisms. 
These conditions impose immense challenge to make constructive suggestions, underscoring the need for representation learning strategies that transcend conventional feature engineering while preserving materials-specific interpretability. 

Learning from small data is currently one of the most intriguing research directions in various fields. Compared to natural language texts and images, scientific data from experimental measurements always falls into the group of small data with lots of physics~\cite{karniadakis_physics_2021}. That is, learning from these data requires sophisticated knowledge conversion to vectorial (or tensorial) material representations. The physical properties derived features are never complete by suffering from the finite and discrete nature. The underlying hidden correlations are not clear enough to be inherited. 
Since generating excessive amount of data is not accomplishable in the short-term, how to effectively represent these data with knowledgeable encodings is the thought-worthy solution.
In addition, deep learning has emerged as the major strategy in various learning applications~\cite{lecun_dl_2015}. The tabular data representation in materials science intrinsically prohibits effective implementation of advanced deep learning algorithms in recommending new materials. This limitation is intrinsic for all kinds of materials.

The first principal question regarding experimental material design is how to directly pick up several elements from the periodic table to synthesize a desired MG with a low critical cooling rate. 
This reminds us of the analogy to the customer-product relationship in marketplaces. For example, for e-commerce like Amazon or Taobao, their websites aim to make effective recommendations for customers with their desired products from the inventories. Recommendation systems with deep learning architectures have emerged as the efficient tool.
We bear in mind that this process does not involve physical synthesis but to uncover the hidden relationship between existing customers and products. This is where the complexity resides in natural science than the marketplaces. The recommended materials will ask experimental fabrication from either element or alloy precursors. Nevertheless, this philosophy indeed provides fresh ideas for materials design.

In this work, we build glass recommendation systems from our proposal of network representations for binary and ternary MGs~\cite{MGnetwork}. This strategy focuses on materials relationships from the perspective of the involved elements and their correlations, rather than treats each composition as independent instances. To learn from these graphs, we also propose novel elemental encodings from the Wikipedia, which is the most precious knowledge library in the world from natural languages. A variety of properties of each element and its links to various Wikipedia pages are encoded in its embeddings. These strategies enable us to build versatile architectures of graph neural networks (GNNs). These models show prominent advantages in predicting MGs in different systems, especially from Transformer-powered GNN~\cite{zhang2024transgnn}. The performance difference arising from different Wikipedia languages also suggest possible knowledge gaps from natural languages and call for deeper knowledge share. The flexibility and versatility of our data-driven machine learning strategy propose a new paradigm in accelerating materials discovery.

\vspace{5mm}
\section*{Results}

\subsection*{Element encoding by Wikipedia}

In general, to build a machine learning protocol, creating effective representations for the input instances is critical. It directly determines the performance of the refined model. This rule applies for every occasion without MG prediction as an outlier. A prototypical example is representing colorful images by pixels. Unfortunately, this becomes extremely difficult for material representation. Standing as a three-dimensional object, materials lack mathematical descriptions, but rather described by their various physical and chemical properties. This is natural as these determine where human beings are able to use them.

%--------------------figure starts-------------------------------
\begin{figure}[!t] %[htp]
\centering
\includegraphics[width = 0.9\textwidth]{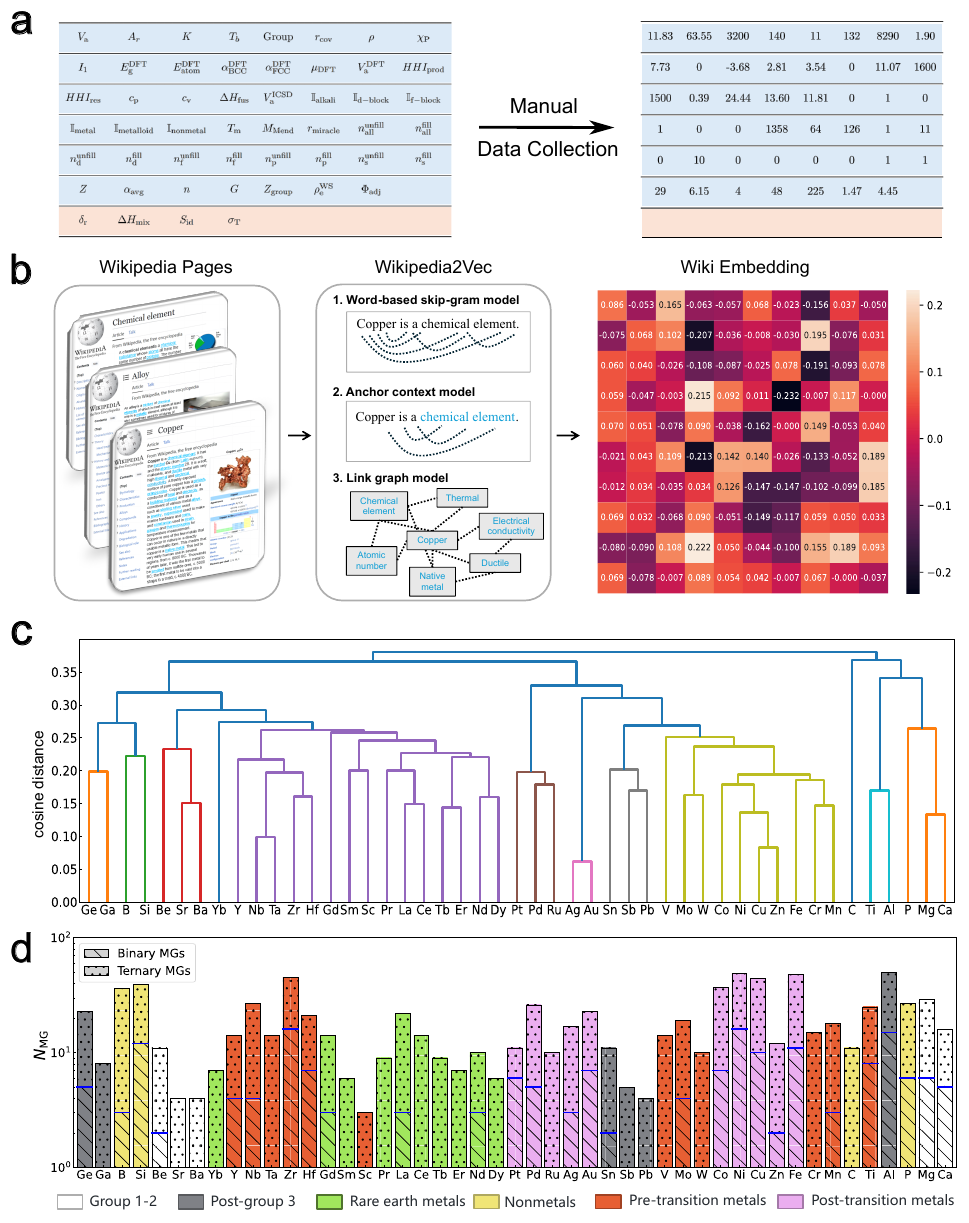}
\caption{
{\bf Material encoding strategies.} 
{\bf a}, Conventional material representation by physical properties of elements and alloys collected manually. There are generally 47 elemental features and 4 alloy features. The right panel shows the data for element Cu as an example.
{\bf b}, Encoding materials by the Wikipedia. The content of a specific Wikipedia page and its link to other pages are processed by the Wikipedia2Vec model. The right panel shows the feature matrix of Cu. 
{\bf c}, The hierarchical clustering dendrogram of element embeddings from Wikipedia.
{\bf d}, Stacked bar plot for the number of MGs that an element is involved in. The upper segment (dotted pattern) represents ternary systems, while the lower segment ($-45^\circ$ diagonal hatches) dictates binary systems. Some elements only show up in ternary MGs. The periodic groups of the corresponding elements color the bars.}
\label{fig1}
\end{figure}
%--------------------figure ends-------------------------------

This fact brings the traditional material representation strategy by vectorizing the quantitative elemental and alloy properties. Figure~\ref{fig1}a shows the typical physical properties to encode a material, such as density $\rho$ of an element and heat-of-mixing $\Delta H_{\rm mix}$ of a mixture. (see \blue{Supplementary Table 1} for the definitions of each symbol) 
Here we show the feature matrix of Cu as an example. These features and their derivative ones from the material composition are  commonly used to make material predictions, including MGs. They are inherently limited in capturing latent chemical interactions. Thus, the question on the completeness of their description for a material naturally arises. In addition, the hidden relationships among these features are unclear. Therefore, this representation method has intrinsic limitations. Together with the tabular data representation for different compositions, the previous studies mainly developed supervised statistical machine learning models, leaving the power of advanced deep learning out of reach. This also suffers risky overfitting from the small dataset.

To circumvent this difficulty, we propose to encode elements from natural language. The comprehensive multi-modal description of an element conveys knowledge to us. To this end, the Wikipedia is, by no means, one of the most important knowledge libraries so far. Its entities are crucial to encode an element or a material directly. 
Inspired from our network representations, we focus on their nodes, i.e. elements, at this moment. The alloys are included in the edge properties.
As illustrated in Fig.~\ref{fig1}b, an open-source API, Wikipedia2Vec~\cite{yamada2020wikipedia2vec}, is used to encode the Wikipedia entities of various elements. This model automatically encode elemental knowledge through three semantic dimensions from Wikipedia pages: 
(1) the conventional skip-gram model learning word-context relationships, 
(2) an anchor context model resolving entity-word co-occurrences, 
and (3) link graph model preserving Wikipedia's hyperlink topology, whose nodes are entities and the edges represent the presence of hyperlinks between the entities. 
As a result, a 100-dimensional ($100d$) embedding is generated for each element in our networks from the English Wikipedia. This embedding generation requires no explicit feature engineering because chemical knowledge emerges organically from Wikipedia's semi-structured scientific discourse. The vector captures latent metallurgical relationships inaccessible to conventional tabular representations (see Fig.~\ref{fig1}a). A heatmap matrix of such an encoding is shown for Cu.

In the following we scrutinize the characteristics of these Wikipedia embeddings and prove their effectiveness in elemental encoding. In Fig.~\ref{fig1}c we plot the hierarchical clustering dendrogram based on the cosine similarity between pair embeddings for the 47 elements depicted in Fig.~\ref{fig1}d that have been used in the material networks. (see Methods) Reasonable closeness within subgroups like Au-Ag, Zr-Hf, and small trees of rare-earth elements are revealed. This hierarchical tree shows the semantic relationship among these elements, showing direct evidence of its effectiveness linked to the periodic table. 
The subgrouping consistency confirms that the embeddings capture fundamental chemical periodicity despite being derived purely from textual knowledge.
This structure also provides advice on element substitution in MG design.

We further show the distribution of the number of MGs by neglecting composition for each element in Fig.~\ref{fig1}d. The data is split into binary systems and ternary systems, respectively. The participation of these elements to the two systems is not even. Generally, ternary systems are better glass-formers than the binary. Therefore, there are more ternary MGs than binary ones from the literature. 
In more detail,  both pre- and post-transition metals dominate the high-yield regions, with Zr forming 16 binary and 72 ternary MGs. Metalloids B and Si occupy strategic positions bridging high-activity transition metals and low-activity post-transition metals, engaging in considerable MGs. This implies their critical role as necessary MG-forming enhancers. Besides, the post-group 3 elements show system-dependent selectivity, with Al forming 15 binary and 93 ternary MGs while Pb only shows up in 4 ternary MGs.
The imbalanced engagement motivates us to build two independent material networks and learn them separately below.

%--------------------figure starts-------------------------------
\begin{figure}[!t] %[htp]
\centering
\includegraphics[width = \textwidth]{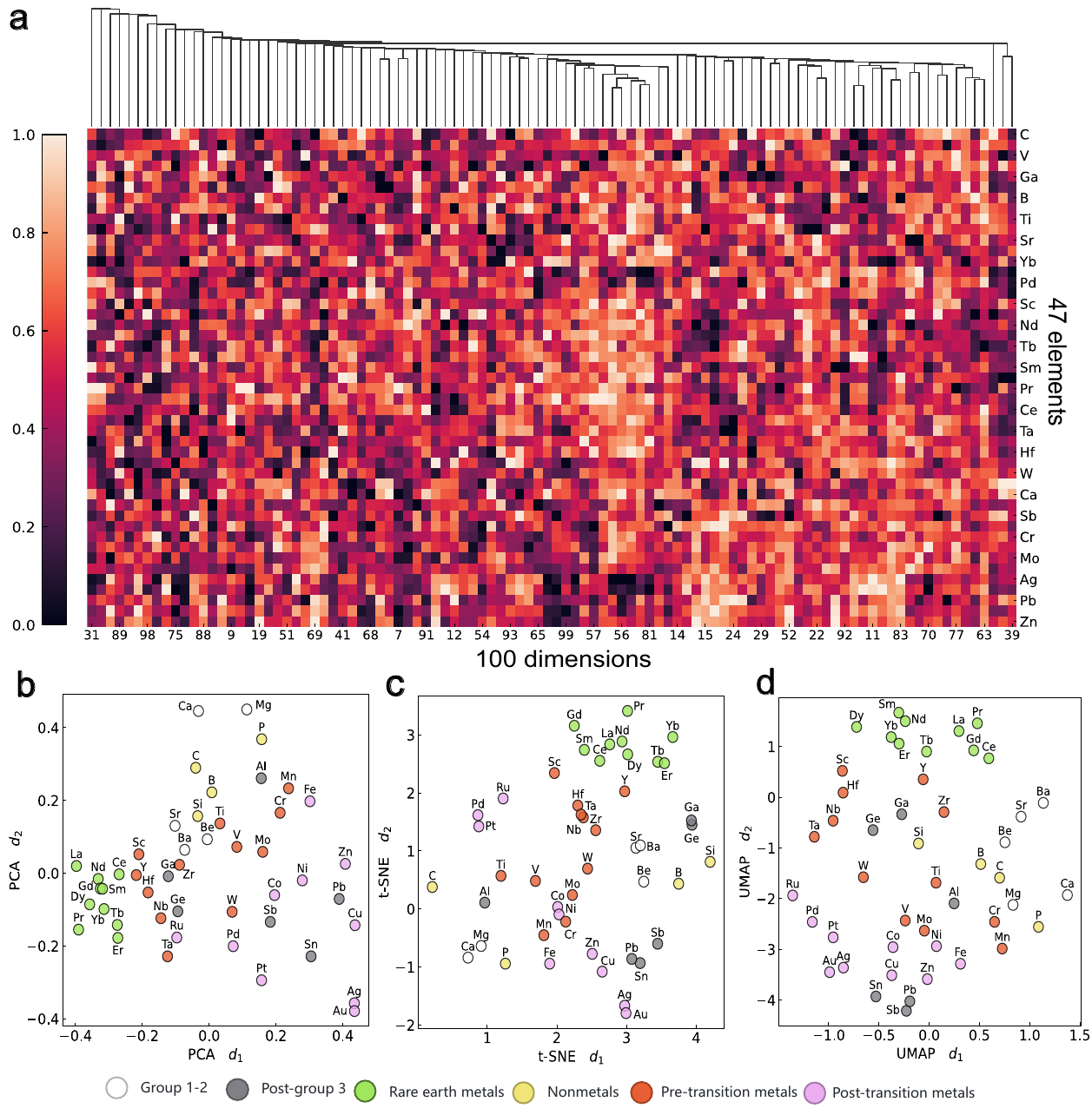}
\caption{
{\bf Correlation analysis of Wikipedia embeddings.} 
{\bf a}, Heatmap of the 100-dimensional Wikipedia embeddings for the 47 elements. The hierarchical clustering dendrogram at the top demonstrates the weak correlation between components of the embeddings. Only part of the labels are shown for clarity.
{\bf b-d}, Dimensionality reduction of element embeddings using PCA ({\bf b}), t-SNE ({\bf c}) and UMAP ({\bf d}), respectively. The elements are illustrated by the top-2 principal components. The data is scattered in all panels, corroborating the importance of each component in the Wikipedia embeddings. A cluster of the rare earth elements is present.}
\label{fig2}
\end{figure}
%--------------------figure ends-------------------------------

After validating the element relationship above, we now focus on analyzing the Wikipedia embeddings themselves. The full matrix of these $100d$ vectors of the 47 elements is visualized in Fig.~\ref{fig2}a. This heatmap demonstrates the avoidance of feature normalization often utilized in statistical learning. On top of it, a hierarchical clustering dendrogram from feature-wise cosine similarity evaluations for the $100d$ features is shown to exclude the subgrouping behavior seen in Fig.~\ref{fig1}c. The tiny tree in middle shows intrinsic similarity for several features from the rare-earth elements. This demonstrates the independence of these individual features.

To further corroborate their independency, we performed dimensionality reduction for direct visualization. Three different popular algorithms are employed, i.e. Principal Component Analysis (PCA), t-Distributed Stochastic Neighbor Embedding (t-SNE), and Uniform Manifold Approximation and Projection (UMAP). Figure~\ref{fig2}b-d show the two-dimensional plots for the 47 elements from the top 2 most important principal components of these algorithms. Remarkably, the universal scattering characteristic is observed from all panels. This demonstrates the intrinsic low redundancy in the Wikipedia embedding space. 
In addition, some interesting subgrouping behavior is also observed, especially for the rare-earth metals. This feature also validates the results in Fig.~\ref{fig1}c. 

From the above analyses of the Wikipedia embeddings for different elements, we learned that they are effective numerical representations in high-dimensional latent space. They capture the intrinsic relationships between elements without referring to their macroscopic physical properties. Meanwhile, the embedding components are non-redundant from each other. These features are of crucial importance to support the construction of a valid machine learning framework.

\subsection*{Graph neural network architectures}

In traditional statistical learning, a complex mapping function is learned from the high-dimensional feature space to reveal the hidden patterns within the input data. A new instance like a material composition is then predicted under the condition that overfitting is necessarily avoided. This learning strategy is born intuitively for tabular data representation. In this process, the feature-to-feature relevance is usually evaluated. Nevertheless, the sample-to-sample relevance is overlooked. 
For instance, in MG discovery, while feature-based methods separately identify Al-Fe and Fe-Nd, sample-level relationships will automatically suggest Al-Nd. By recognizing that samples containing Al-Fe and Fe-Nd share contextual similarities, the system infers untested Al-Nd configurations through their implicit sample neighborhood associations – a connection invisible to isolated feature comparisons.
This intrinsic limitation has enormous impact for model prediction because it limits the capacity to model higher-order compositional synergies critical for GFA. The sequential treatment on the sample-to-sample relationships is the power engine of the successful Transformer model, and further to the large language models.

To overcome these difficulties, we initiate advanced graph neural networks (GNNs) on the basis of the material networks for binary and ternary MGs we proposed previously~\cite{MGnetwork}. Within this framework, we encode the nodes with the derived Wikipedia embeddings and the entities (link or triangle) dictate specific a MG system. This graph topology explicitly encodes multi-element interactions. Three popular algorithms are considered to build the glass recommendation system. They are Graph Convolutional Networks (GCN)~\cite{tomas2017gcn}, Neural Graph Collaborative Filtering (NGCF)~\cite{wang2019neural}, and Transformer-based GNNs (TransGNN)\cite{zhang2024transgnn}, respectively. These architectures learn context-aware embeddings through structured message passing. (see details of these algorithms in Methods) 

%--------------------figure starts-------------------------------
\begin{figure}[!t] %[htp]
\centering
\includegraphics[width = \textwidth]{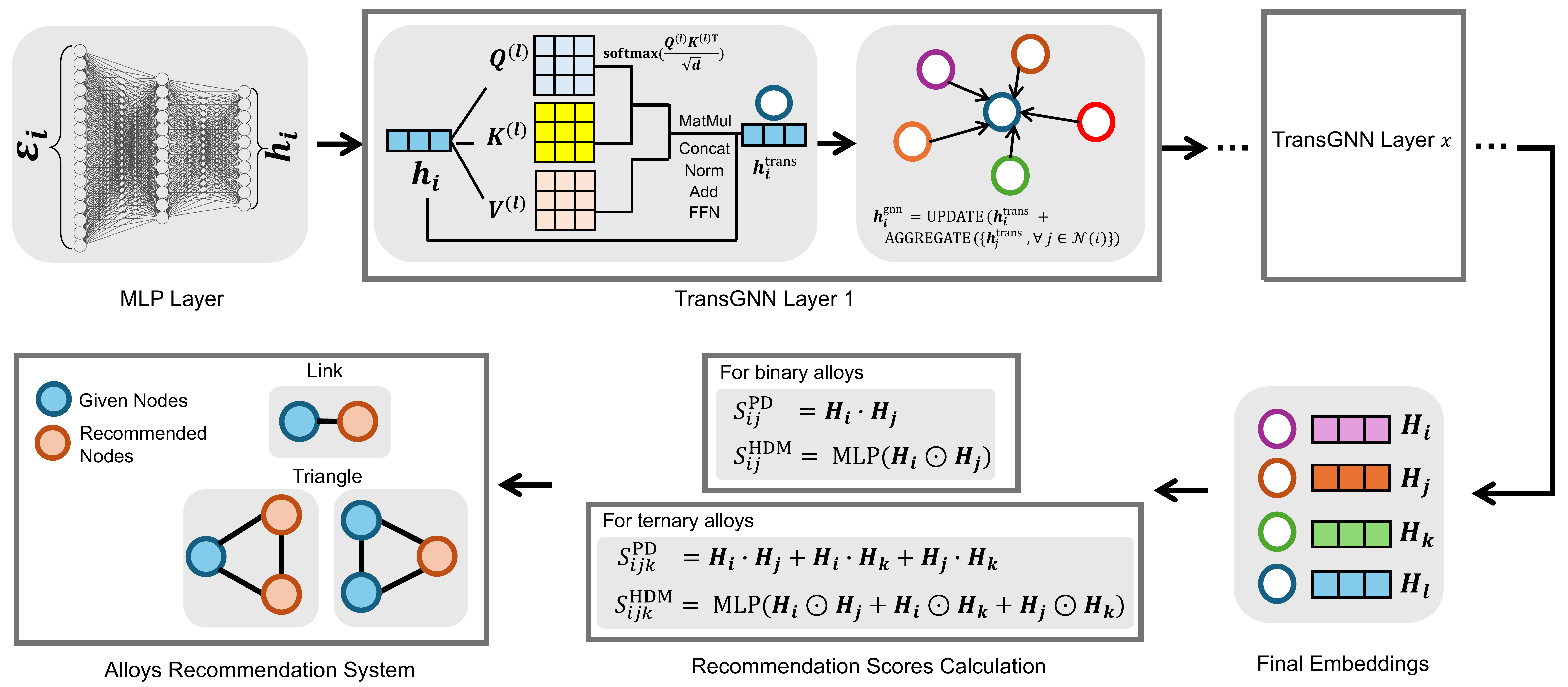}
\caption{
{\bf Graph-based recommendation system for MGs.}
The input elemental Wikipedia embeddings are mapped to node representations by MLP layers. By embeddings post-processing (e.g. by Transformer) and advanced message passing, different GNN models (e.g. TransGNN) are designed to generated the final embeddings. A recommendation system is thus built to score the appearance of a link (binary system) or a triangle (ternary system) in the material networks. (PD: inner product; HDM: Hadamard product)}
\label{fig3}
\end{figure}
%--------------------figure ends-------------------------------

We then frame MG discovery as a graph recommendation system to suggest probable glass-forming element pairs (for binary systems) or completes ternary compositions (via element/pair recommendations), transcending the limitations of Euclidean distance metrics and supervised classification paradigms. 

Figure~\ref{fig3} visualizes the workflow of the glass recommendation system with TransGNN as the example algorithm. 
In brief, the initial $100d$ Wikipedia embeddings are firstly processed by multi-layer perceptron (MLP) to project elemental features into a shared latent space to compress semantic knowledge for graph-based operations. Subsequent TransGNN layers integrate the Transformer self-attention mechanisms with GNN message passing, enabling elements to dynamically update their embeddings by aggregating contextual information from neighboring nodes. This dual mechanism captures both local topological dependencies (via graph aggregation) and global relational patterns (via attention weights). Following successive TransGNN layers, the final elemental representations are learned for scoring.

To make glass recommendations, we use two schemes, the inner product (PD) and the element-wise Hadamard product (HDM), to score the suggested entity. The calculation formulas are shown in the Fig.~\ref{fig3}. Basically, PD is more straightforward but HDM is learnable from MLP. Therefore, HDM is more versatile with enhanced nonlinear characteristics, especially for high-order entity prediction.
For all tasks, the choice between PD and HDM scoring is data-driven, with the former suitable for capturing linear dependencies in well-characterized interaction spaces and the latter excelling in complex, non-linear compositional landscapes. As illustrated in the schematic, this modular design supports both node-centric (predicting missing elements) and pair-centric (predicting missing pairs) recommendation modes, providing flexibility to address diverse discovery scenarios. 

The model training process is designed to optimize the recommendation performance while preventing overfitting. (see Methods)
We refer positive samples as those input alloys fabricated in experiments. The non-amorphous forming element pairs and invalid triplets containing at least one element from positive samples but failing amorphous formation criteria are taken as negative samples for binary and ternary networks. 
The training objective combines ranking-aware supervision with regularization constraints.
A pairwise ranking loss with L2 regularization is defined as:
\begin{equation}
    \mathcal{L} = -\sum^{N_\text{pos}}_{j} \log\left[\sigma(S_j^{+} - S_j^{-}) + \epsilon\right] + 
    \lambda \sum_{i} \|\theta_i\|^2,
\end{equation}
where $N_{\rm pos}$ is the number of positive samples, $S_j^{+}$ represents the recommendation score (see Methods) for positive sample $j$, and $S_j^{-}$  represents the recommendation score of the randomly selected negative sample corresponding to the postivie sample $j$.  
The function $\sigma$ is the sigmoid function, and $\epsilon$ is an extremely small positive constant added for numerical stability. $\lambda$ is the L2 regularization coefficient and $\theta_i$ is the $i$-th trainable parameter.

We employ the stratified 5-fold cross-validation to evaluate model performance. Each fold iteration computes two metrics: Recall@K and Normalized Discounted Cumulative Gain (NDCG@K), where $K$ denotes the number of top-ranked recommendations. (see Methods) 
We focus on $K=10$ in the current study. The first metric considers how many correct predictions are made in the top-$K$ group, serving currently as the major rule. While the second accounts for the order of these predictions. The purpose is to make sure the highly recommended samples are correct at the top of the list. 

The optimal model configuration is selected based on maximum Recall@K achieved on the test sets across all folds. The final model is subsequently trained on the complete dataset using the validated hyperparameters. The involved hyperparameters have been optimized by intensive grid search (see Methods).

%--------------------figure starts-------------------------------
\begin{figure}[!t] %[htp]
\centering
\includegraphics[width = \textwidth]{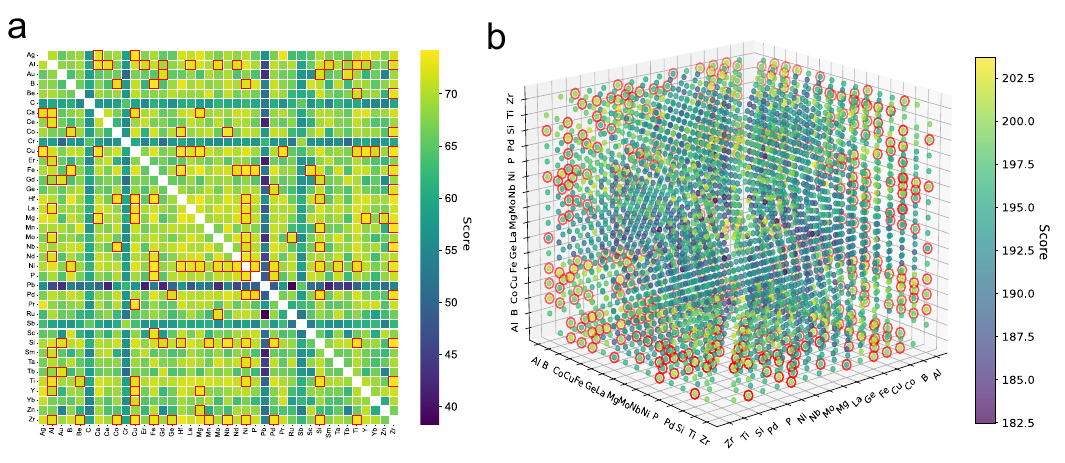}
\caption{{\bf Visualization of the recommendation scores for MG prediction. }
{\bf a}, Two-dimensional heatmap of the scores for binary MG recommendation by TransGNN. The red squares mark those validated by experiments previously.
{\bf b}, Three-dimensional scatter plot of the scores for ternary MG recommendation by TransGNN. The red circles are exemplified experimental validations from 16 elements for better clarity.}
\label{fig4}
\end{figure}
%--------------------figure ends-------------------------------

\section*{Glass recommendation systems}
In our previous study~\cite{MGnetwork}, we built the binary network with 38 nodes and 94 edges from the experimentally fabricated binary MGs. For the ternary network, it is constructed from 47 elements and 352 triangles, also from the experimentally synthesized ternary MGs. (see Methods for more details, and network visualization below) 
Note that the construction entities are different between the two networks. 
Based on the above discussion, we build three glass recommendation systems:
(1) ``B2B" to predict binary MGs from the binary network, 
(2) ``T2T" to predict ternary MGs from the ternary network, and 
(3) ``B2T" to predict the ternary MGs from the binary network.
Before diving into the comprehensive analyses for each system, we show the general performance of B2B and T2T by TransGNN in Figure~\ref{fig4}. The diverse distributions of the recommendation scores indicate the broad range of GFA for these alloys. More importantly, the higher score suggests better GFA, which proves the robustness of TransGNN by overlapping the existing MGs (red squares in Fig.~\ref{fig4}a and red circles in Fig.~\ref{fig4}b) with the high-score recommendations. 
This spatial alignment sets the stage for quantitative performance evaluation across all three recommendation tasks.

For these different recommendation systems, we explore the roles of GNN architectures, scoring schemes, and the input embeddings in determining the model performance. 
In brief, the GNN architectures are GCN, NGCF and TransGNN. PD and HDM are the scoring scheme candidates. For the input embeddings, other than the English (ENG) Wikipedia discussed above, we also generate elemental embeddings from the Wikipedia entities in another 10 languages in the similar way. They cover English (ENG), Chinese (CHN), Japanese (JAPAN), French (FREN), German (GER), Spanish (SPAN), Italian (ITAL), Portuguese (PORT), Russian (RUSS), Polish (POL), and Dutch (DUT).

\subsection*{1. B2B recommendation system}

\begin{figure}[!t] %[htp]
\centering
\includegraphics[width = \textwidth]{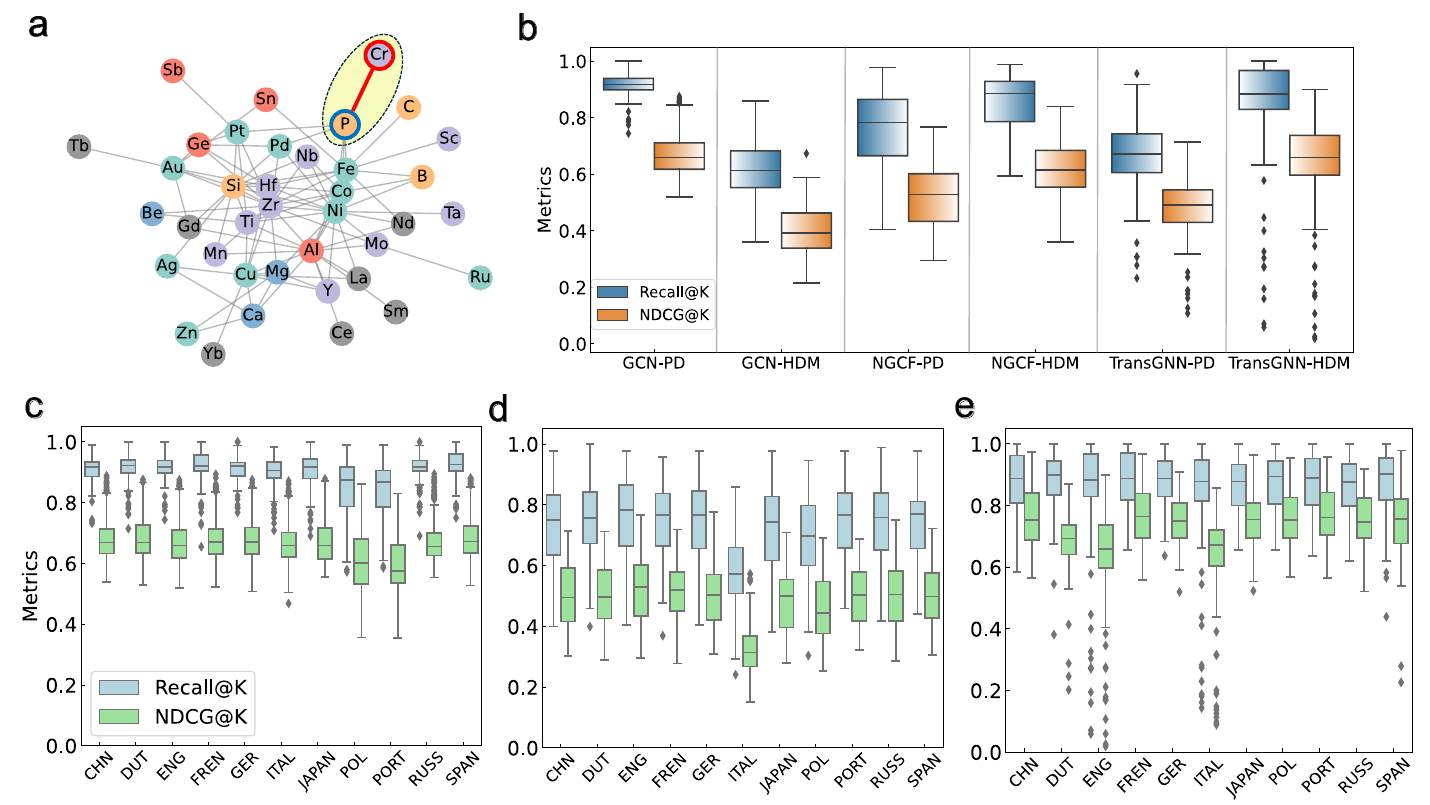}
\caption{
{\bf Performance of the B2B recommendation system.}
{\bf a}, Planar visualization of the binary network with a link recommendation highlighted.
{\bf b}, Model performance evaluation for three GNN architectures over the English Wikipedia embeddings.
{\bf c-e}, Metrics for multi-lingual recommendation systems with GCN-PD ({\bf c}), NGCF-PD ({\bf d}), and TransGNN-HDM ({\bf e}).
}
\label{fig5}
\end{figure}

Figure~\ref{fig5}a shows the schematic of recommending a link from the binary network. In principle, B2B trained from the binary network takes a single element as input and predicts compatible partners.
After training, the final embedding of each node captures contextual interactions within the graph. For a given query element, recommendation scores for links with all other nodes are computed to prioritize probable glass-forming partners. The yellow eclipse in Fig.~\ref{fig5}a highlights such exemplar predictions.

%--------------------figure starts-------------------------------
%--------------------figure ends-------------------------------

Figure~\ref{fig5}b quantifies the performance across 30 training trials based on Recall@K and NDCG@K, comparing GCN, NGCF, and TransGNN frameworks with both PD and HDM scoring strategies. Notably, the GCN-PD architecture achieves the highest metrics, with Recall@K being $0.915 \pm 0.041$ and NDCG@K being $0.683 \pm 0.085$, outperforming other learning methods. This suggests that for the relatively simple network, models with less complexity can behave better than more complicated ones. It also provides better explainability. Nevertheless, if considering the HDM scheme, TransGNN performs the best, rendering larger performance fluctuations.

Figure~\ref{fig5}c-e present box plots of Recall@K and NDCG@K across 30 independent runs using Wikipedia embeddings derived from multi-lingual embeddings, evaluating the robustness of our framework to language-specific knowledge encoding. 
Notably, all GNN architectures, GCN-PD (c), NGCF-PD (d), and TransGNN-HDM (e), exhibit small performance variance across languages, especially GCN-PD. This stability indicates that Wikipedia embeddings, despite being trained on language-specific scientific discourse, capture universal elemental relationships critical for glass-forming prediction, with language-specific nuances having negligible impact on model performance. This also indicates that the basic knowledge has been included in the Wikipedia of diverse languages. But the varying median and variance remind the knowledge gap.
More remarkably, GCN-PD maintains the highest median Recall@K and NDCG@K across all language settings, reinforcing its efficiency in leveraging linear similarity metrics for binary systems. GCN-PD’s consistent superiority highlights its optimal balance of simplicity and effectiveness for the B2B recommendation system, where pairwise interactions dominate and complex nonlinear transformations offer diminishing returns.

\subsection*{2. T2T recommendation system}

%--------------------figure starts-------------------------------
\begin{figure}[!t] %[htp]
\centering
\includegraphics[width = \textwidth]{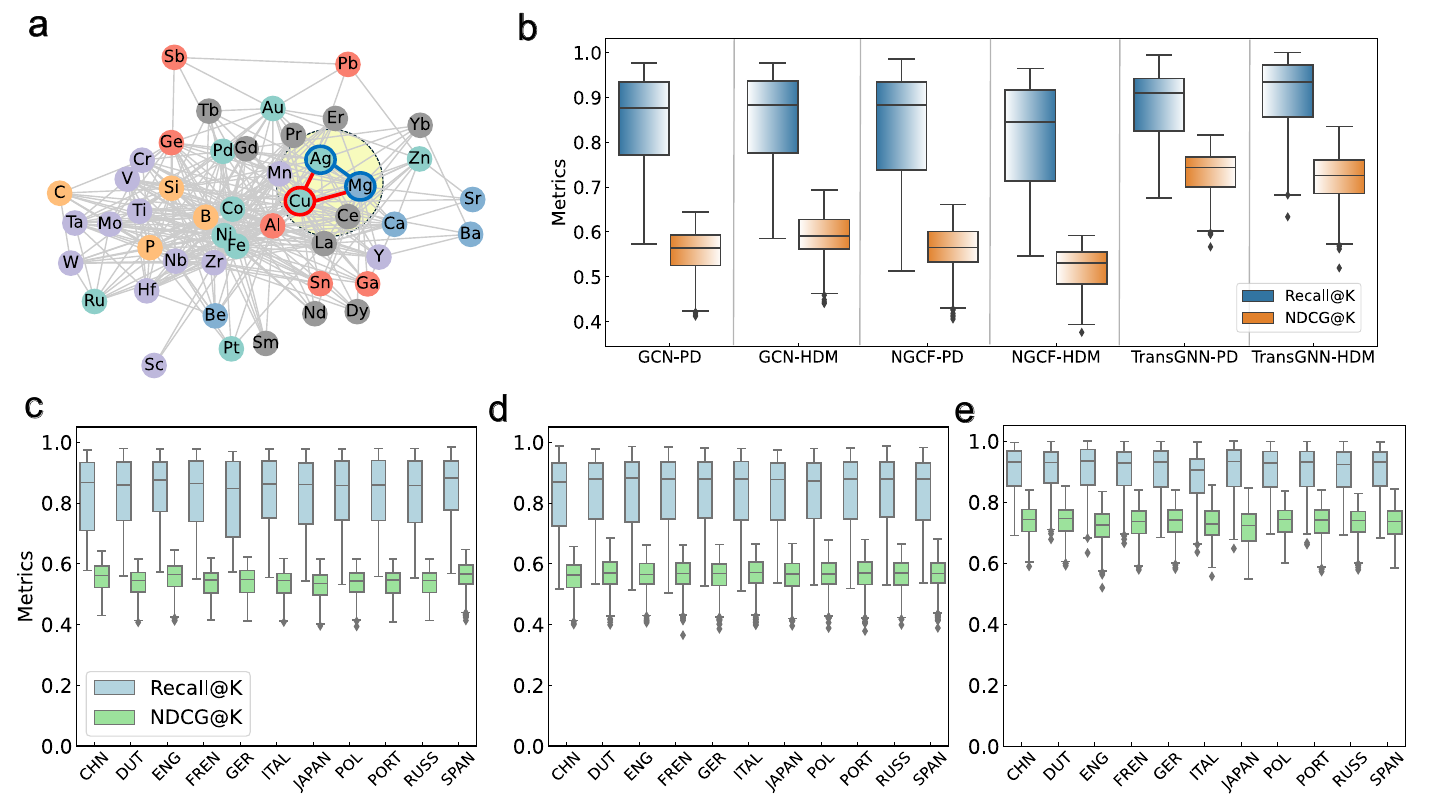}
\caption{
{\bf Performance of the T2T recommendation system.}
{\bf a}, Two-dimensional visualization of the ternary network with a triangle recommendation highlighted.
{\bf b}, Model performance evaluation for three GNN architectures over the English Wikipedia embeddings.
{\bf c-e}, Metrics for multi-lingual recommendation systems with GCN-PD ({\bf c}), NGCF-PD ({\bf d}), and TransGNN-HDM ({\bf e}.)
}
\label{fig6}
\end{figure}
%--------------------figure ends-------------------------------

In the ternary network, the recommendation system supports two complementary tasks: (i) given an element, recommending a compatible pair to complete the ternary system; or (ii) given a known pair, suggesting the third element to form a ternary system. 
While the main text focuses on demonstrating scenario (ii), comprehensive results for task (i) are systematically presented in \blue{Supplementary Fig. 1}. Notably, conclusions from scenario (ii) align consistently with those derived from task (i). 
Figure~\ref{fig6}a shows the two-dimensional projection of the ternary network, with a case of (ii) illustrated. A query input of the known pair Ag-Mg suggests Cu as the top-10 recommended third element.
For both scenarios, the recommendation score is derived from the aggregated similarity of all three pairwise combinations, ensuring holistic evaluation of multi-element synergies. This approach explicitly models the cooperative nature of ternary glass formation, where three-body interactions are crucial to dictate GFA.

Figure~\ref{fig6}b-e mirror the analysis strategy of those in Fig.~\ref{fig5} by evaluating Recall@K and NDCG@K across 30 training trials for GCN, NGCF, and TransGNN architectures with PD or HDM scoring. While all models exhibit excellent performance, TransGNN emerges as the best model. The HDM scoring exhibits higher Recall@K than the PD result, even though their NDCG@K values are more or less comparative. 
This indicates a more complicated model is naturally necessary to model complex networks. The advantage of TransGNN can be attributed to the enhanced message passing by the attention mechanism to capture long-distance hidden relationships.
The box plots in Fig~\ref{fig6}c-e resemble the findings from Fig~\ref{fig5}: negligible performance variance across the Wikipedia embeddings over 11 languages. This consistency reinforces that the fundamental chemical principles governing ternary glass formation are robustly captured by the elemental embeddings from the Wikipedia knowledge library, irrespective of the natural language.

\subsection*{3. B2T recommendation system}
Even though the binary network is constructed from the reported binary MGs, the network itself automatically generates high-order entities, such as triangle, rectangle, and so on. Here we consider the triangle entity to bridge the two networks. This also marks the capability of predicting high-order objects from simpler networks, which is out of the capability of the traditional statistical learning from the tabular data representation. We will report more comprehensive analysis on the deep connections between the binary network and the ternary network in a separate study.

In principle, identifying a good binary glass-former is an important step to design the ternary systems. A typical example is by adding Al to CuZr, the critical diameter of the amorphous rod can increase from 2 mm for Cu$_{50}$Zr$_{50}$ to 8 mm for Cu$_{46}$Zr$_{46}$Al$_8$~\cite{tang_binary_2004,Yu_excellent_2005,zhang_glass_2007}. 
Fantastic mechanical properties may also emerge~\cite{das_work_2005}.
Furthermore, this can be linked to the famous minor alloying effect in the MG field. That is, adding a small amount of a third element to a master alloy can greatly enhance its GFA. The mystery is still far from clearly understood. The network representation may provide fresh clues to guide such material design from purely topology analysis.

%--------------------figure starts-------------------------------
\begin{figure}[!t] %[htp]
\centering
\includegraphics[width = \textwidth]{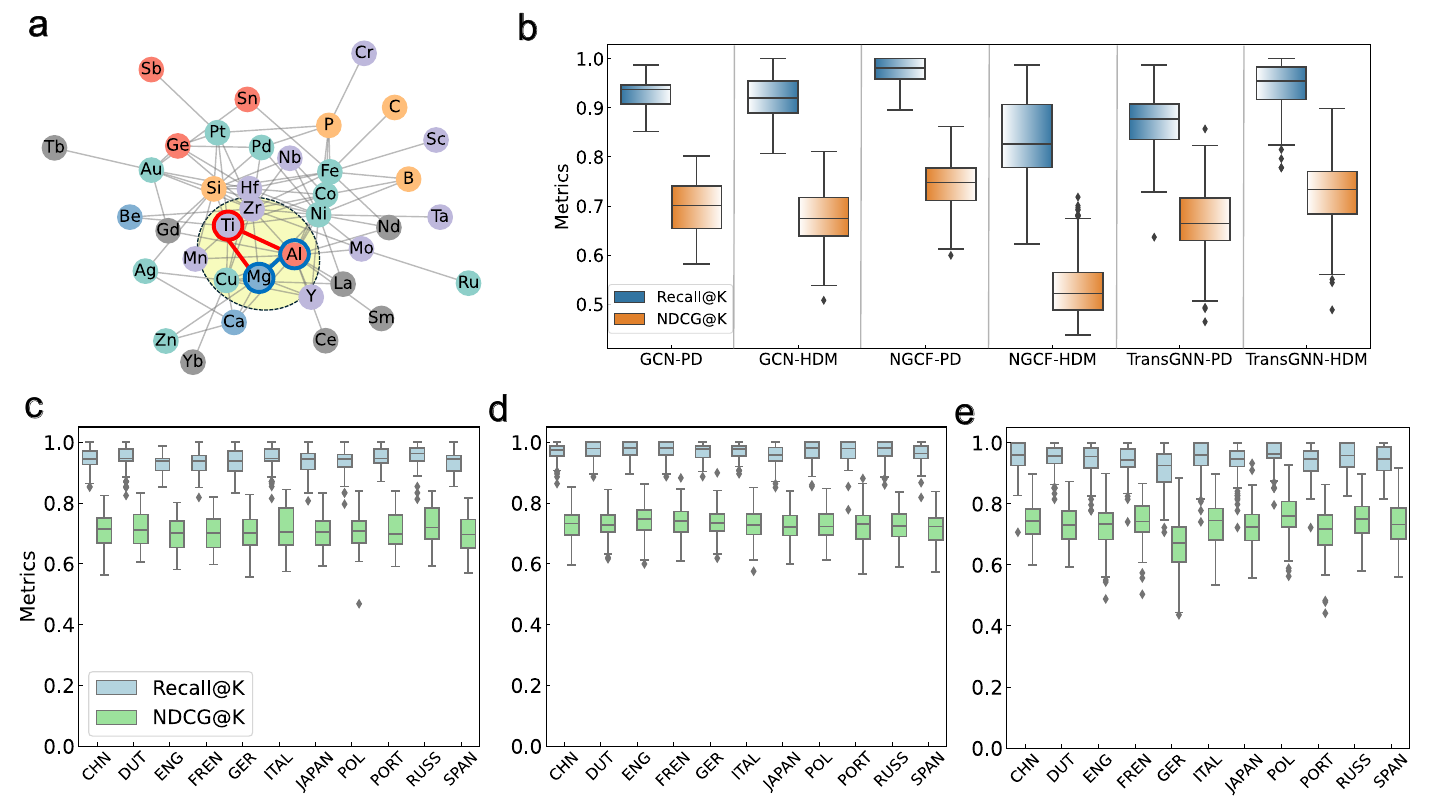}
\caption{
{\bf Performance of the B2T recommendation system.}
{\bf a}, Planar visualization of the binary network with a triangle recommendation highlighted.
{\bf b}, Model performance evaluation for three GNN architectures over the English Wikipedia embeddings.
{\bf c-e}, Metrics for multi-lingual recommendation systems with GCN-PD ({\bf c}), NGCF-PD ({\bf d}), and TransGNN-HDM ({\bf e}).
}
\label{fig7}
\end{figure}
%--------------------figure ends-------------------------------

Here we focus on building the B2T recommendation systems to predict possible ternary MGs from the binary network. The schematic is depicted in Fig.~\ref{fig7}a. A representative case is shown, where a B2T system successfully recommends Ti as the third element for the Al-Mg pair, predicting the Al-Mg-Ti ternary MG. This suggests the knowledge share in the formation of binary and ternary MGs, which is unknown currently.

Figure~\ref{fig7}b compares the model performance among GCN, NGCF, and TransGNN with combinations of PD or HDM. It is interesting to find that NGCF is the best model, with much higher Recall@K than the others. 
The high accuracy may stem from NGCF’s high-order connectivity modeling and hierarchical embedding propagation. By capturing multi-hop relations in the binary network (e.g., indirect element interactions via shared neighbors), NGCF infers implicit ternary compatibility from synergistic binary patterns. Simultaneously, embedding propagation aggregates neighborhood features across graph layers, preserving direct binary correlations while propagating latent ternary collaboration signals, thereby mitigating data sparsity and encoding physicochemical constraints critical to multi-component alloy formation. 
This observation along with the B2B and T2T systems emphasize the architecture-specific model for different material design problems. 
Consistently, Fig.~\ref{fig7}c-e marks minimal performance variation for the Wikipedia embeddings from different languages. The robust cross-network prediction capability highlights the framework’s utility in accelerating discovery for under-explored multi-element systems, especially in the case where direct data is scarce but transferable knowledge is abundant. The ability to bridge binary and ternary scales, which is supported by semantically rich embeddings and graph-based transfer learning, establishes a generalizable paradigm for multi-component material discovery, offering unprecedented efficiency in navigating the enormous material space.

\section*{Discussion}

In summary, we build glass recommendation systems with versatile graph neural network architectures based on our proposed network representations for MGs.  The GFA information is inherently carried by the network topology. More significantly, the nodes are encoded from the Wikipedia by a language model. These context-aware embeddings leverage semantic patterns in elemental Wikipedia pages, which captures the latent materials knowledge embedded in scientific discourse without relying on predefined physical descriptors. 
This approach recommends probable glass-forming combinations based on topological proximity and embedding similarity, bypassing the constraints of supervised classification limited by tabular data patterns. The comparison between the PD and HDM scoring scheme suggests the effectiveness of HDM in predicting high-order entities in networks for future study.
Our strategy will reshape the way of intelligent materials design by natural language processings.

By comparing the model performance, we identify the problem-oriented learning strategy design. For example, GCN with a simple framework performs the best to predict binary MGs from the binary network. TransGNN is the best model to predict ternary MGs from the more complex network. The NGCF architecture outperforms in transfer learning to bridge the binary and ternary networks. Building a universal model is very challenging but highly desired, probably with the ensemble strategy.

Finally, the small performance variances of the Wikipedia embeddings from the 11 languages convey important information. It confirms the alignment of the fundamental knowledge of chemical elements in the Wikipedia library. The encoded elemental semantics are universal and robust to linguistic context. This language-agnostic knowledge extraction opens a new route for materials discovery frameworks. 
Another intriguing point is that the current Wikipedia libraries all capture the most basic knowledge about elements for MG design, but not advanced enough with domain expertise. This possibly explains the inevitable gap of Recall@K from the perfect. More sophisticated learning methodologies are highly demanded to further improve the elemental embeddings to accelerate smart material design, especially for MGs.

Our findings pioneer the decoupling of descriptor engineering from domain heuristics through language model-derived representations. The inherent compatibility with emerging multi-modal architectures---particularly when integrated with large language models (LLM) for automated knowledge extraction---suggests transformative potential for next-generation materials design platforms.
The established recommendation system with top-notch deep learning techniques serves as a powerful paradigm for data-driven MG discovery, offers transformative advances in semantic feature engineering and cross-scale recommendation. Looking forward, integrating LLMs to automate scientific knowledge extraction could unlock a new era of autonomous materials design. By bridging linguistic semantics, graph analytics, and recommendation system, this work charts a course for next-generation materials informatics, where complexity is met with scalable, interpretable, and knowledge-driven machine intelligence.

%\newpage

%-------------------references for main text------------------
%\bibliographystyle{naturemag_noURL}
%\bibliography{mainref}

\vspace{1cm}
\begin{thebibliography}{10}
\expandafter\ifx\csname url\endcsname\relax
  \def\url#1{\texttt{#1}}\fi
\expandafter\ifx\csname urlprefix\endcsname\relax\def\urlprefix{URL }\fi
\providecommand{\bibinfo}[2]{#2}
\providecommand{\eprint}[2][]{\url{#2}}

\bibitem{inoue_recent_2011}
\bibinfo{author}{Inoue, A.} \& \bibinfo{author}{Takeuchi, A.}
\newblock \bibinfo{title}{Recent development and application products of bulk
  glassy alloys}.
\newblock \emph{\bibinfo{journal}{Acta Mater.}} \textbf{\bibinfo{volume}{59}},
  \bibinfo{pages}{2243--2267} (\bibinfo{year}{2011}).

\bibitem{wang_roadmap_2025}
\bibinfo{author}{Wang, W.} \emph{et~al.}
\newblock \bibinfo{title}{Metallic glass roadmap}.
\newblock \emph{\bibinfo{journal}{Mater. Futures}}  (\bibinfo{year}{2025}).

\bibitem{Eliaz_biomaterials_2019}
\bibinfo{author}{Eliaz, N.}
\newblock \bibinfo{title}{Corrosion of metallic biomaterials: A review}.
\newblock \emph{\bibinfo{journal}{Mater.}} \textbf{\bibinfo{volume}{12}}
  (\bibinfo{year}{2019}).

\bibitem{Demetriou_damage_2011}
\bibinfo{author}{Demetriou, M.~D.}, \bibinfo{author}{Launey, M.~E.},
  \bibinfo{author}{Garrett, G.}, \bibinfo{author}{Schramm, J.~P.},
  \bibinfo{author}{Hofmann, D.~C.}, \bibinfo{author}{Johnson, W.~L.} \&
  \bibinfo{author}{Ritchie, R.~O.}
\newblock \bibinfo{title}{A damage-tolerant glass}.
\newblock \emph{\bibinfo{journal}{Nat. Mater.}} \textbf{\bibinfo{volume}{10}},
  \bibinfo{pages}{123--128} (\bibinfo{year}{2011}).

\bibitem{inoue_stabilization_2000}
\bibinfo{author}{Inoue, A.}
\newblock \bibinfo{title}{Stabilization of metallic supercooled liquid and bulk
  amorphous alloys}.
\newblock \emph{\bibinfo{journal}{Acta Mater.}} \textbf{\bibinfo{volume}{48}},
  \bibinfo{pages}{279--306} (\bibinfo{year}{2000}).

\bibitem{inoue_preparation_1997}
\bibinfo{author}{Inoue, A.}, \bibinfo{author}{Nishiyama, N.} \&
  \bibinfo{author}{Kimura, H.}
\newblock \bibinfo{title}{Preparation and thermal stability of bulk amorphous
  {Pd40Cu30Ni10P20} alloy cylinder of 72 mm in diameter}.
\newblock \emph{\bibinfo{journal}{Mater. Trans. JIM}}
  \textbf{\bibinfo{volume}{38}}, \bibinfo{pages}{179--183}
  (\bibinfo{year}{1997}).

\bibitem{ward_machine_2018}
\bibinfo{author}{Ward, L.}, \bibinfo{author}{O'Keeffe, S.~C.},
  \bibinfo{author}{Stevick, J.}, \bibinfo{author}{Jelbert, G.~R.},
  \bibinfo{author}{Aykol, M.} \& \bibinfo{author}{Wolverton, C.}
\newblock \bibinfo{title}{A machine learning approach for engineering bulk
  metallic glass alloys}.
\newblock \emph{\bibinfo{journal}{Acta Mater.}} \textbf{\bibinfo{volume}{159}},
  \bibinfo{pages}{102--111} (\bibinfo{year}{2018}).

\bibitem{liu2020machine}
\bibinfo{author}{Liu, X.}, \bibinfo{author}{Li, X.}, \bibinfo{author}{He, Q.},
  \bibinfo{author}{Liang, D.}, \bibinfo{author}{Zhou, Z.}, \bibinfo{author}{Ma,
  J.}, \bibinfo{author}{Yang, Y.} \& \bibinfo{author}{Shen, J.}
\newblock \bibinfo{title}{Machine learning-based glass formation prediction in
  multicomponent alloys}.
\newblock \emph{\bibinfo{journal}{Acta Mater.}} \textbf{\bibinfo{volume}{201}},
  \bibinfo{pages}{182--190} (\bibinfo{year}{2020}).

\bibitem{turnbull_under_1969}
\bibinfo{author}{Turnbull, D.}
\newblock \bibinfo{title}{Under what conditions can a glass be formed?}
\newblock \emph{\bibinfo{journal}{Contemp. Phys.}}
  \textbf{\bibinfo{volume}{10}}, \bibinfo{pages}{473--488}
  (\bibinfo{year}{1969}).

\bibitem{johnson_bulk_1999}
\bibinfo{author}{Johnson, W.~L.}
\newblock \bibinfo{title}{Bulk glass-forming metallic alloys: Science and
  technology}.
\newblock \emph{\bibinfo{journal}{MRS Bull.}} \textbf{\bibinfo{volume}{24}},
  \bibinfo{pages}{42--56} (\bibinfo{year}{1999}).

\bibitem{ding_combinatorial_2014}
\bibinfo{author}{Ding, S.}, \bibinfo{author}{Liu, Y.}, \bibinfo{author}{Li,
  Y.}, \bibinfo{author}{Liu, Z.}, \bibinfo{author}{Sohn, S.},
  \bibinfo{author}{Walker, F.~J.} \& \bibinfo{author}{Schroers, J.}
\newblock \bibinfo{title}{Combinatorial development of bulk metallic glasses}.
\newblock \emph{\bibinfo{journal}{Nat. Mater.}} \textbf{\bibinfo{volume}{13}},
  \bibinfo{pages}{494--500} (\bibinfo{year}{2014}).

\bibitem{li_high_2019}
\bibinfo{author}{Li, M.-X.}, \bibinfo{author}{Zhao, S.-F.},
  \bibinfo{author}{Lu, Z.}, \bibinfo{author}{Hirata, A.}, \bibinfo{author}{Wen,
  P.}, \bibinfo{author}{Bai, H.-Y.}, \bibinfo{author}{Chen, M.},
  \bibinfo{author}{Schroers, J.}, \bibinfo{author}{Liu, Y.} \&
  \bibinfo{author}{Wang, W.-H.}
\newblock \bibinfo{title}{High-temperature bulk metallic glasses developed by
  combinatorial methods}.
\newblock \emph{\bibinfo{journal}{Nature}} \textbf{\bibinfo{volume}{569}},
  \bibinfo{pages}{99--103} (\bibinfo{year}{2019}).

\bibitem{sun2017machine}
\bibinfo{author}{Sun, Y.-T.}, \bibinfo{author}{Bai, H.-Y.},
  \bibinfo{author}{Li, M.-Z.} \& \bibinfo{author}{Wang, W.-H.}
\newblock \bibinfo{title}{Machine learning approach for prediction and
  understanding of glass-forming ability}.
\newblock \emph{\bibinfo{journal}{J. Phys. Chem. Lett.}}
  \textbf{\bibinfo{volume}{8}}, \bibinfo{pages}{3434--3439}
  (\bibinfo{year}{2017}).

\bibitem{ren_accelerated_2018}
\bibinfo{author}{Ren, F.}, \bibinfo{author}{Ward, L.},
  \bibinfo{author}{Williams, T.}, \bibinfo{author}{Laws, K.~J.},
  \bibinfo{author}{Wolverton, C.}, \bibinfo{author}{Hattrick-Simpers, J.} \&
  \bibinfo{author}{Mehta, A.}
\newblock \bibinfo{title}{Accelerated discovery of metallic glasses through
  iteration of machine learning and high-throughput experiments}.
\newblock \emph{\bibinfo{journal}{Sci. Adv.}} \textbf{\bibinfo{volume}{4}},
  \bibinfo{pages}{eaaq1566} (\bibinfo{year}{2018}).

\bibitem{wen2019machine}
\bibinfo{author}{Wen, C.}, \bibinfo{author}{Zhang, Y.}, \bibinfo{author}{Wang,
  C.}, \bibinfo{author}{Xue, D.}, \bibinfo{author}{Bai, Y.},
  \bibinfo{author}{Antonov, S.}, \bibinfo{author}{Dai, L.},
  \bibinfo{author}{Lookman, T.} \& \bibinfo{author}{Su, Y.}
\newblock \bibinfo{title}{Machine learning assisted design of high entropy
  alloys with desired property}.
\newblock \emph{\bibinfo{journal}{Acta Mater.}} \textbf{\bibinfo{volume}{170}},
  \bibinfo{pages}{109--117} (\bibinfo{year}{2019}).

\bibitem{lu2020interpretable}
\bibinfo{author}{Lu, Z.} \emph{et~al.}
\newblock \bibinfo{title}{Interpretable machine-learning strategy for
  soft-magnetic property and thermal stability in {Fe}-based metallic glasses}.
\newblock \emph{\bibinfo{journal}{Npj Comput. Mater.}}
  \textbf{\bibinfo{volume}{6}}, \bibinfo{pages}{187} (\bibinfo{year}{2020}).

\bibitem{xiong2020machine}
\bibinfo{author}{Xiong, J.}, \bibinfo{author}{Shi, S.-Q.} \&
  \bibinfo{author}{Zhang, T.-Y.}
\newblock \bibinfo{title}{A machine-learning approach to predicting and
  understanding the properties of amorphous metallic alloys}.
\newblock \emph{\bibinfo{journal}{Mater. Des.}} \textbf{\bibinfo{volume}{187}},
  \bibinfo{pages}{108378} (\bibinfo{year}{2020}).

\bibitem{peng2021determination}
\bibinfo{author}{Peng, L.}, \bibinfo{author}{Long, Z.} \&
  \bibinfo{author}{Zhao, M.}
\newblock \bibinfo{title}{Determination of glass forming ability of bulk
  metallic glasses based on machine learning}.
\newblock \emph{\bibinfo{journal}{Comput. Mater. Sci.}}
  \textbf{\bibinfo{volume}{195}}, \bibinfo{pages}{110480}
  (\bibinfo{year}{2021}).

\bibitem{batra_emerging_2021}
\bibinfo{author}{Batra, R.}, \bibinfo{author}{Song, L.} \&
  \bibinfo{author}{Ramprasad, R.}
\newblock \bibinfo{title}{Emerging materials intelligence ecosystems propelled
  by machine learning}.
\newblock \emph{\bibinfo{journal}{Nat. Rev. Mater.}}
  \textbf{\bibinfo{volume}{6}}, \bibinfo{pages}{655--678}
  (\bibinfo{year}{2021}).

\bibitem{afflerbach_machine_2022}
\bibinfo{author}{Afflerbach, B.~T.} \emph{et~al.}
\newblock \bibinfo{title}{Machine learning prediction of the critical cooling
  rate for metallic glasses from expanded datasets and elemental features}.
\newblock \emph{\bibinfo{journal}{Chem. Mater.}} \textbf{\bibinfo{volume}{34}},
  \bibinfo{pages}{2945--2954} (\bibinfo{year}{2022}).

\bibitem{li_data_2022}
\bibinfo{author}{Li, M.-X.}, \bibinfo{author}{Sun, Y.-T.},
  \bibinfo{author}{Wang, C.}, \bibinfo{author}{Hu, L.-W.},
  \bibinfo{author}{Sohn, S.}, \bibinfo{author}{Schroers, J.},
  \bibinfo{author}{Wang, W.-H.} \& \bibinfo{author}{Liu, Y.-H.}
\newblock \bibinfo{title}{Data-driven discovery of a universal indicator for
  metallic glass forming ability}.
\newblock \emph{\bibinfo{journal}{Nat. Mater.}} \textbf{\bibinfo{volume}{21}},
  \bibinfo{pages}{165--172} (\bibinfo{year}{2022}).

\bibitem{forrest2023evolutionary}
\bibinfo{author}{Forrest, R.~M.} \& \bibinfo{author}{Greer, A.~L.}
\newblock \bibinfo{title}{Evolutionary design of machine-learning-predicted
  bulk metallic glasses}.
\newblock \emph{\bibinfo{journal}{Digit. Discov.}}
  \textbf{\bibinfo{volume}{2}}, \bibinfo{pages}{202--218}
  (\bibinfo{year}{2023}).

\bibitem{merchant_scaling_2023}
\bibinfo{author}{Merchant, A.}, \bibinfo{author}{Batzner, S.},
  \bibinfo{author}{Schoenholz, S.~S.}, \bibinfo{author}{Aykol, M.},
  \bibinfo{author}{Cheon, G.} \& \bibinfo{author}{Cubuk, E.~D.}
\newblock \bibinfo{title}{Scaling deep learning for materials discovery}.
\newblock \emph{\bibinfo{journal}{Nature}} \textbf{\bibinfo{volume}{624}},
  \bibinfo{pages}{80--85} (\bibinfo{year}{2023}).

\bibitem{liu2023machine}
\bibinfo{author}{Liu, G.}, \bibinfo{author}{Sohn, S.}, \bibinfo{author}{Kube,
  S.~A.}, \bibinfo{author}{Raj, A.}, \bibinfo{author}{Mertz, A.},
  \bibinfo{author}{Nawano, A.}, \bibinfo{author}{Gilbert, A.},
  \bibinfo{author}{Shattuck, M.~D.}, \bibinfo{author}{O'Hern, C.~S.} \&
  \bibinfo{author}{Schroers, J.}
\newblock \bibinfo{title}{Machine learning versus human learning in predicting
  glass-forming ability of metallic glasses}.
\newblock \emph{\bibinfo{journal}{Acta Mater.}} \textbf{\bibinfo{volume}{243}},
  \bibinfo{pages}{118497} (\bibinfo{year}{2023}).

\bibitem{liu2024effective}
\bibinfo{author}{Liu, G.}, \bibinfo{author}{Sohn, S.}, \bibinfo{author}{O'Hern,
  C.~S.}, \bibinfo{author}{Gilbert, A.~C.} \& \bibinfo{author}{Schroers, J.}
\newblock \bibinfo{title}{Effective subgrouping enhances machine learning
  prediction in complex materials science phenomena: Inoue's subgrouping in
  discovering bulk metallic glasses}.
\newblock \emph{\bibinfo{journal}{Acta Mater.}} \textbf{\bibinfo{volume}{265}},
  \bibinfo{pages}{119590} (\bibinfo{year}{2024}).

\bibitem{hu_data_2023}
\bibinfo{author}{Hu, Y.-C.} \& \bibinfo{author}{Tian, J.}
\newblock \bibinfo{title}{Data-driven prediction of the glass-forming ability
  of modeled alloys by supervised machine learning}.
\newblock \emph{\bibinfo{journal}{J. Mater. Inf.}}
  \textbf{\bibinfo{volume}{3}}, \bibinfo{pages}{1} (\bibinfo{year}{2023}).

\bibitem{xie_catalogue_2025}
\bibinfo{author}{Xie, W.}, \bibinfo{author}{Li, M.}, \bibinfo{author}{Sun, Y.},
  \bibinfo{author}{Wang, C.}, \bibinfo{author}{Hu, L.} \& \bibinfo{author}{Liu,
  Y.}
\newblock \bibinfo{title}{A catalogue of metallic glass-forming alloy systems}.
\newblock \emph{\bibinfo{journal}{Materialia}} \textbf{\bibinfo{volume}{39}},
  \bibinfo{pages}{102375} (\bibinfo{year}{2025}).

\bibitem{karniadakis_physics_2021}
\bibinfo{author}{Karniadakis, G.~E.}, \bibinfo{author}{Kevrekidis, I.~G.},
  \bibinfo{author}{Lu, L.}, \bibinfo{author}{Perdikaris, P.},
  \bibinfo{author}{Wang, S.} \& \bibinfo{author}{Yang, L.}
\newblock \bibinfo{title}{Physics-informed machine learning}.
\newblock \emph{\bibinfo{journal}{Nat. Rev. Phys.}}
  \textbf{\bibinfo{volume}{3}}, \bibinfo{pages}{422--440}
  (\bibinfo{year}{2021}).

\bibitem{lecun_dl_2015}
\bibinfo{author}{LeCun, Y.}, \bibinfo{author}{Bengio, Y.} \&
  \bibinfo{author}{Hinton, G.}
\newblock \bibinfo{title}{Deep learning}.
\newblock \emph{\bibinfo{journal}{Nature}} \textbf{\bibinfo{volume}{521}},
  \bibinfo{pages}{436--444} (\bibinfo{year}{2015}).

\bibitem{MGnetwork}
\bibinfo{author}{Zhang, S.}, \bibinfo{author}{Tian, J.}, \bibinfo{author}{Liu,
  S.}, \bibinfo{author}{Zhang, H.}, \bibinfo{author}{Bai, H.},
  \bibinfo{author}{Wang, W.-H.} \& \bibinfo{author}{Hu, Y.-C.}
\newblock \bibinfo{title}{Data-driven material networks for intelligent matter
  design}.
\newblock \emph{\bibinfo{journal}{unpublished}}  (\bibinfo{year}{2025}).

\bibitem{zhang2024transgnn}
\bibinfo{author}{Zhang, P.}, \bibinfo{author}{Yan, Y.}, \bibinfo{author}{Zhang,
  X.}, \bibinfo{author}{Li, C.}, \bibinfo{author}{Wang, S.},
  \bibinfo{author}{Huang, F.} \& \bibinfo{author}{Kim, S.}
\newblock \bibinfo{title}{Transgnn: Harnessing the collaborative power of
  transformers and graph neural networks for recommender systems}.
\newblock In \emph{\bibinfo{booktitle}{Proc. 47th Int. ACM SIGIR Conf. Res.
  Dev. Inf. Retr.}}, \bibinfo{pages}{1285--1295} (\bibinfo{year}{2024}).

\bibitem{yamada2020wikipedia2vec}
\bibinfo{author}{Yamada, I.}, \bibinfo{author}{Asai, A.},
  \bibinfo{author}{Sakuma, J.}, \bibinfo{author}{Shindo, H.},
  \bibinfo{author}{Takeda, H.}, \bibinfo{author}{Takefuji, Y.} \&
  \bibinfo{author}{Matsumoto, Y.}
\newblock \bibinfo{title}{{W}ikipedia2{V}ec: An efficient toolkit for learning
  and visualizing the embeddings of words and entities from {W}ikipedia}.
\newblock In \emph{\bibinfo{booktitle}{Proceedings of the 2020 Conference on
  Empirical Methods in Natural Language Processing: System Demonstrations}},
  \bibinfo{pages}{23--30} (\bibinfo{publisher}{Association for Computational
  Linguistics}, \bibinfo{year}{2020}).

\bibitem{tomas2017gcn}
\bibinfo{author}{Kipf, T.~N.} \& \bibinfo{author}{Welling, M.}
\newblock \bibinfo{title}{Semi-supervised classification with graph
  convolutional networks}.
\newblock In \emph{\bibinfo{booktitle}{5th International Conference on Learning
  Representations, {ICLR} 2017, Toulon, France, April 24-26, 2017, Conference
  Track Proceedings}} (\bibinfo{publisher}{OpenReview.net},
  \bibinfo{year}{2017}).

\bibitem{wang2019neural}
\bibinfo{author}{Wang, X.}, \bibinfo{author}{He, X.}, \bibinfo{author}{Wang,
  M.}, \bibinfo{author}{Feng, F.} \& \bibinfo{author}{Chua, T.-S.}
\newblock \bibinfo{title}{Neural graph collaborative filtering}.
\newblock In \emph{\bibinfo{booktitle}{Proceedings of the 42nd international
  ACM SIGIR conference on Research and development in Information Retrieval}},
  \bibinfo{pages}{165--174} (\bibinfo{year}{2019}).

\bibitem{tang_binary_2004}
\bibinfo{author}{Tang, M.-B.}, \bibinfo{author}{Zhao, D.-Q.},
  \bibinfo{author}{Pan, M.-X.} \& \bibinfo{author}{Wang, W.-H.}
\newblock \bibinfo{title}{Binary {Cu}-{Zr} bulk metallic glasses}.
\newblock \emph{\bibinfo{journal}{Chin. Phys. Lett.}}
  \textbf{\bibinfo{volume}{21}}, \bibinfo{pages}{901--903}
  (\bibinfo{year}{2004}).

\bibitem{Yu_excellent_2005}
\bibinfo{author}{Yu, P.}, \bibinfo{author}{Bai, H.}, \bibinfo{author}{Tang, M.}
  \& \bibinfo{author}{Wang, W.}
\newblock \bibinfo{title}{Excellent glass-forming ability in simple
  {Cu}50{Zr}50-based alloys}.
\newblock \emph{\bibinfo{journal}{J. Non-Cryst. Solids}}
  \textbf{\bibinfo{volume}{351}}, \bibinfo{pages}{1328--1332}
  (\bibinfo{year}{2005}).

\bibitem{zhang_glass_2007}
\bibinfo{author}{Zhang, Q.}, \bibinfo{author}{Zhang, W.}, \bibinfo{author}{Xie,
  G.} \& \bibinfo{author}{Inoue, A.}
\newblock \bibinfo{title}{Glass-forming ability and mechanical properties of
  the ternary {C}u-{Z}r-{A}l and quaternary {C}u-{Z}r-{A}l-{A}g bulk metallic
  glasses}.
\newblock \emph{\bibinfo{journal}{Mater. Trans.}}
  \textbf{\bibinfo{volume}{48}}, \bibinfo{pages}{1626--1630}
  (\bibinfo{year}{2007}).

\bibitem{das_work_2005}
\bibinfo{author}{Das, J.}, \bibinfo{author}{Tang, M.}, \bibinfo{author}{Kim,
  K.}, \bibinfo{author}{Theissmann, R.}, \bibinfo{author}{Baier, F.},
  \bibinfo{author}{Wang, W.} \& \bibinfo{author}{Eckert, J.}
\newblock \bibinfo{title}{“{Work}-hardenable” ductile bulk metallic glass}.
\newblock \emph{\bibinfo{journal}{Phys. Rev. Lett.}}
  \textbf{\bibinfo{volume}{94}}, \bibinfo{pages}{205501}
  (\bibinfo{year}{2005}).

\bibitem{li_how_2017}
\bibinfo{author}{Li, Y.}, \bibinfo{author}{Zhao, S.}, \bibinfo{author}{Liu,
  Y.}, \bibinfo{author}{Gong, P.} \& \bibinfo{author}{Schroers, J.}
\newblock \bibinfo{title}{How many bulk metallic glasses are there?}
\newblock \emph{\bibinfo{journal}{ACS Comb. Sci.}}
  \textbf{\bibinfo{volume}{19}}, \bibinfo{pages}{687--693}
  (\bibinfo{year}{2017}).

\bibitem{schultz_exploration_2021}
\bibinfo{author}{Schultz, L.~E.}, \bibinfo{author}{Afflerbach, B.},
  \bibinfo{author}{Francis, C.}, \bibinfo{author}{Voyles, P.~M.},
  \bibinfo{author}{Szlufarska, I.} \& \bibinfo{author}{Morgan, D.}
\newblock \bibinfo{title}{Exploration of characteristic temperature
  contributions to metallic glass forming ability}.
\newblock \emph{\bibinfo{journal}{Comput. Mater. Sci.}}
  \textbf{\bibinfo{volume}{196}}, \bibinfo{pages}{110494}
  (\bibinfo{year}{2021}).

\bibitem{kingma2014adam}
\bibinfo{author}{Kingma, D.~P.} \& \bibinfo{author}{Ba, J.}
\newblock \bibinfo{title}{Adam: A method for stochastic optimization}.
\newblock \emph{\bibinfo{journal}{arXiv preprint arXiv:1412.6980}}
  (\bibinfo{year}{2014}).

\end{thebibliography}

\vspace{1cm}
%----------------------METHODS-----------------------------
%\clearpage
\section*{Methods}
\subsection*{Material networks for binary and ternary MGs}
The experimental datasets of amorphous alloys were systematically curated from authoritative materials databases and peer-reviewed literature~\cite{ren_accelerated_2018, ward_machine_2018, li_how_2017,schultz_exploration_2021}.  Following our previous study~\cite{MGnetwork}, we abstract chemical elements as nodes in two distinct networks, i.e., binary and ternary MG networks. 
The compiled dataset contains 94 validated binary amorphous systems and 352 ternary systems, serving as positive samples for model training. The non-amorphous forming element pairs and invalid triplets containing at least one element from positive samples but failing amorphous formation criteria are taken as negative samples for binary and ternary networks, respectively.

\subsection*{Elemental Wikipedia embedding generation}
Element representations were derived through multilingual semantic analysis using Wikipedia2Vec~\cite{yamada2020wikipedia2vec}, an open-source toolkit that generates joint embeddings of words and Wikipedia entities which integrates the conventional skip-gram model with two extended architectures—the anchor context model and the link graph model—to jointly embed words and entities into a unified 100-dimensional vector space.
For each chemical element, corresponding Wikipedia pages were extracted across 11 languages: English (ENG), Chinese (CHN), Japanese (JAPAN), French (FREN), German (GER), Spanish (SPAN), Italian (ITAL), Portuguese (PORT), Russian (RUSS), Polish (POL), and Dutch (DUT). The embedding $\mathcal{E}$ is defined as $\mathcal{E} = \{ e_\ell \in \mathbb{R}^{100} \mid \ell \in L \}$, where $e_\ell$ represents the 100-dimensional embedding vector for a given element in language $\ell$, and $L = \{ \ell_1, \dots, \ell_{11} \}$ denotes the set of supported languages. These embeddings serve as the initial representations for elemental nodes in the subsequent graph-based framework.

\subsection*{Hierarchical clustering dendrogram}
To investigate the structural relationships between learned element representations, we performed agglomerative hierarchical clustering on the embedding vectors using cosine similarity as the affinity metric. The embeddings were first normalized to  ensure scale invariance, as required by cosine-based similarity measures. Pairwise similarity scores between embeddings were calculated using the cosine of the angle between vectors, which quantifies their directional alignment in high-dimensional space. These similarity scores were then converted to pairwise distances by subtracting each similarity value from 1, thereby transforming the similarity range into a distance metric.
The clustering procedure utilized Ward’s linkage criterion, which iteratively merges clusters to minimize the total within-cluster variance. This method promotes the formation of cohesive spherical clusters in the embedding space. The hierarchical merging process was continued until all elements were grouped under a single root node, generating a complete dendrogram structure. %Analyses were conducted using the SciPy library’s hierarchical clustering tools, with dendrogram visualization generated via matplotlib.

\subsection*{GNN algorithms}
GCN aggregates node features through spectral-based convolution. Given an adjacency matrix $\mathbf{A} \in \mathbb{R}^{N \times N}$, GCN constructs a normalized adjacency matrix 
\begin{equation}
    \hat{\mathbf{A}} = \mathbf{D}^{-1/2} (\mathbf{A} + \mathbf{I}) \mathbf{D}^{-1/2},
\end{equation}
where $\mathbf{D}$ is the degree matrix and $\mathbf{I}$ is the identity matrix. Feature updates are computed as $\mathbf{H} = \hat{\mathbf{A}} \mathbf{X} \mathbf{W}$, with $\mathbf{X} \in \mathbb{R}^{N \times d_{\text{in}}}$ as the input features, $\mathbf{W} \in \mathbb{R}^{d_{\text{in}} \times d_{\text{out}}}$ as the trainable weight matrix, and $\mathbf{H}$ as the output feature matrix. 
The activation function ReLU and the dropout are allplied after this step.

NGCF integrates feature transformation and interaction modeling. The features of node $i$ are directly transformed via $\mathbf{h}_i = \mathbf{W}_1 \mathbf{x}_i$, where $\mathbf{W}_1$ is a weight matrix, and $\mathbf{x}_i$ is the input features of node $i$. Interactions between neighboring nodes are modeled as 
\begin{equation}
    \mathbf{m}_{ij} = \frac{\mathbf{x}_i \odot \mathbf{x}_j}{\sqrt{d_i^{\text{deg}} d_j^{\text{deg}}}}, 
\end{equation}
with $d_i^{\text{deg}}$ and $d_j^{\text{deg}}$ representing node degrees for node $i$ and $j$ and $\odot$ denoting the Hadamard product. The aggregated interaction feature for node $i$ is then computed as 
\begin{equation}
    \mathbf{h}_i^{\text{int}} = \frac{1}{|\mathcal{N}(i)|} \sum_{j \in \mathcal{N}(i)} \mathbf{W}_2 \mathbf{m}_{ij}, 
\end{equation}
where $\mathcal{N}(i)$ is the neighbor set of node $i$. The final node representation combines both components: $\mathbf{h}_i^{\text{final}} = \mathbf{h}_i + \mathbf{h}_i^{\text{int}}$.
Then the $\text{LeakyReLU}$ activation function is applied, followed by dropout after the transformation to prevent overfitting.

The TransGNN module intergrates both the Transformer layer and the GNN layer,  processing node features through self-attention and graph-based aggregation. The input  features $\mathbf{h}_i$ of node $i$ are first passed through a Transformer layer, which captures the global relationships between nodes through attention. The output of the Transformer layer is then passed to the GNN layer for further refinement. For head $l$ of a transformer layer, the query ($\mathbf{Q}^{(l)}$), key ($\mathbf{K}^{(l)}$), and value ($\mathbf{V}^{(l)}$) matrices are derived from the input feature $\mathbf{h}_i$ of node $i$  as $\mathbf{Q}^{(l)} = \mathbf{h}_i \mathbf{W}_Q^{(l)}$, $\mathbf{K}^{(l)} = \mathbf{h}_i \mathbf{W}_K^{(l)}$, and $\mathbf{V}^{(l)} = \mathbf{h}_i \mathbf{W}_V^{(l)}$, where $\mathbf{W}_Q^{(l)}$, $\mathbf{W}_K^{(l)}$, and $\mathbf{W}_V^{(l)}$ are learnable weight matrices. The context vector for node $i$ in the $l$-th attention head is computed as:
\begin{equation}
    {\mathbf{C}}_i^{(l)} = \text{softmax}\left( \frac{{\mathbf{Q}}^{(l)} {\mathbf{K}}^{(l)\top}}{\sqrt{d}} \right) {\mathbf{V}}^{(l)},
\end{equation}
where $d$ is the key vector dimensionality.
Multi-head attention extends this by applying multiple independent attention heads, concatenating their outputs as 
\begin{equation}
    \mathbf{C}_i = \text{Concat}(\mathbf{C}_i^{(1)}, \mathbf{C}_i^{(2)}, \dots, \mathbf{C}_i^{(\text{num\_heads})}),
\end{equation}
where $\text{num\_heads}$ is set to 4 in this work.
This concatenated context vector is then passed through a linear transformation with weight matrix $\mathbf{W}_O$:
\begin{equation}
    \mathbf{C}_i^{\text{final}} = \mathbf{C}_i \mathbf{W}_O.
\end{equation}
The output of the Transformer layer is computed by adding a residual connection from the input node features and applying dropout, followed by layer normalization:
\begin{equation}
    \mathbf{h}_i^{'} = \text{LayerNorm}(\mathbf{h}_i + \mathbf{C}_i^{\text{final}}).
\end{equation}
In addition, a feedforward network (FFN) is applied to the output of the attention mechanism:
\begin{equation}
    \mathbf{h}_i^{\text{ffn}} = \text{FFN}(\mathbf{h}_i^{'}) = \text{ReLU}(\mathbf{W}_{\rm f1} \mathbf{h}_i^{'}) \mathbf{W}_{\rm f2},
\end{equation}
where $\mathbf{W}_{\rm f1}$ and $\mathbf{W}_{\rm f2}$ are learnable weight matrices in the feedforward network.
Finally, a residual connection is applied to the output of the FFN, followed by layer normalization:
\begin{equation}
    \mathbf{h}_i^{\text{trans}} = \text{LayerNorm}(\mathbf{h}_i^{'} + \mathbf{h}_i^{\text{ffn}}).
\end{equation}

The output of the Transformer layer, $\mathbf{h}_i^{\text{trans}}$, is then passed to the GNN layer.
In the GNN layer, for each node $i$, the message aggregation from its neighbors is defined as:
\begin{equation}
    \mathbf{h}_i^{\text{agg}} = \frac{1}{|\mathcal{N}(i)|} \sum_{j \in \mathcal{N}(i)} \mathbf{W}_2 \mathbf{h}^{\rm trans}_j,
\end{equation}
where $\mathbf{W}_2$ is a learnable weight matrix for the neighbor features, and $\mathcal{N}(i)$ is the set of neighbors of node $i$. After computing the messages from the neighbors, dropout is applied to prevent overfitting.
Once the aggregated messages from neighbors are computed, the next step is to combine the original node feature $\mathbf{h}_i^{\rm trans}$ with the aggregated messages $\mathbf{h}_i^{\text{agg}}$. This is done through concatenation:
\begin{equation}
    \mathbf{h}_i^{\text{combined}} = \text{Concat}(\mathbf{h}_i^{\rm trans}, \mathbf{h}_i^{\text{agg}}).
\end{equation}
The concatenated feature vector is then passed through a linear layer $\mathbf{W}_1$ and ReLU activation, followed by layer normalization:
\begin{equation}
    \mathbf{h}_i^{\text{gnn}} = \text{LayerNorm}(\text{ReLU}(\mathbf{W}_1 \mathbf{h}_i^{\text{combined}})).
\end{equation}
Finally, dropout is applied again to the output of the GNN layer to prevent overfitting.
The final output of the GNN layer, $\mathbf{h}_i^{\text{gnn}}$, represents the updated node features after aggregation and transformation.

\subsection*{Recommendation score definition}
The recommendation system employs scoring strategies tailored to predict binary alloys (B2B) or ternary alloys (B2T, T2T). Two operators—--inner product (PD) and Hadamard product (HDM)—--are used to compute affinity scores. For binary alloys, the inner product score is 
$S^{\text{PD}}_{ij} = \mathbf{H}_i \cdot \mathbf{H}_j$, 
while the Hadamard product score is 
$S^{\text{HDM}}_{ij} = \text{MLP}(\mathbf{H}_i \odot \mathbf{H}_j)$. 
For ternary alloys, the inner product score extends to 
$S^{\text{PD}}_{ijk} = \mathbf{H}_i \cdot\mathbf{H}_j + \mathbf{H}_i\cdot \mathbf{H}_k + \mathbf{H}_j\cdot \mathbf{H}_k$, 
and the Hadamard product score is $S^{\text{HDM}}_{ijk} = \text{MLP}(\mathbf{H}_i \odot \mathbf{H}_j + \mathbf{H}_i \odot \mathbf{H}_k + \mathbf{H}_j \odot \mathbf{H}_k)$. 
These scores guide the prediction of promising alloy compositions.

\subsection*{Model optimization}

The model parameters are updated using the Adam optimizer~\cite{kingma2014adam}.
Critical hyperparameters were determined through grid search:
\begin{itemize}
    \item L2 regularization coefficient $ \lambda \in \{0.01, 0.001, 0.0001\} $
    \item Learning rate $ \in \{ 10^{-6},10^{-5},10^{-4},10^{-3},10^{-2},10^{-1}  \}$
    \item Dropout rate $ \in \{0,0.1,0.2, 0.3, 0.4, 0.5,0.6,0.7, 0.8\} $ for both MLP and Transformer layers
    \item Number of GNN layers $ \in \{1, 2, 3\} $
\end{itemize}

The optimal model configuration is selected based on maximum Recall@K achieved on the test sets across all folds. The final model is subsequently trained on the complete dataset using the optimized hyperparameters.
To mitigate overfitting and ensure robust generalization, a stratified 5-fold cross-validation framework was implemented during training, with each fold allocating 80\% of data for training and 20\% for holdout testing.

\subsection*{Evaluation metrics}

We define two typical metrics for recommendation systems. The first is 
Recall@K to quantify the fraction of relevant (positive) samples retrieved within the top-$K$ predictions:
\begin{equation}
        \text{Recall@K} = \frac{|\mathcal{R} \cap \mathcal{P}_K|}{\min(|\mathcal{R}|, K)},
\end{equation}
where $\mathcal{R}$ denotes the set of relevant items and $\mathcal{P}_K$ represents the top-$K$ predicted items.
The second is 
NDCG@K to assess ranking quality through position-weighted relevance scoring:
    \begin{align}
        \text{DCG@K} &= \sum_{i=1}^K \frac{\text{rel}(i)}{\log_2(i + 1)}, \\
        \text{IDCG@K} &= \sum_{i=1}^{\min(K, |\mathcal{R}|)} \frac{1}{\log_2(i + 1)}, \\
        \text{NDCG@K} &= \frac{\text{DCG@K}}{\text{IDCG@K}},
    \end{align}
where $\text{rel}(i) \in \{0,1\}$ indicates binary relevance at position $i$. The metric ranges within $[0,1]$, with 1 representing perfect ranking alignment. Because of the large fluctuations of the number of positive samples especially in the binary network, we made several attempts for $K \in [5, 10, 15, 20]$. The comparison suggests $K=10$ as a reasonable choice.

%-------------------references------------------
%\section*{References}
%\bibliographystyle{naturemag_noURL}

%----------------additional part---------------------------
\section *{Acknowledgments}
This work is supported by the National Natural Science Foundation of China (Grant No. 52471178).
The support from the Chinese Academy of Sciences (XDB0510000) is also acknowledged.

\section *{Author contributions:}
Y.C.H. conceived and supervised the project. Y.C.H., S.Y.Z., S.L.L. and J.T. prepared the datasets and material networks. Y.C.H. generated element embeddings from the Wikipedia. K.O. built the graph neural network models and machine learning workflows under the joint supervision of Y.C.H. and H.T. S.Y.Z. performed large-scale calculations for hyperparameter optimization with the support of S.L.L. All the authors contributed to the data analysis and results discussion. S.Y.Z. and Y.C.H. wrote the manuscript.

\subsection *{Competing interests:}
The authors declare no competing interests.

\subsection *{Correspondence.}
Correspondence and requests for materials should be addressed to Y.C.H. or H.T.

%\subsection *{Additional information.}
%{\bf Supplementary information} is available for this paper at this url.

%--------------------------extended data figures------------------------------------
%\renewcommand{\figurename}{{\bf Extended Data Fig.}}
%\setcounter{figure}{0}
%
%\clearpage
%\begin{figure}[htp]
%\centering
%%\includegraphics[width = 0.5\textwidth]{figs1.pdf}
%\caption{figure caption}
%\label{figs1}
%\end{figure}

\balance
\end{document}